\documentclass[journal]{IEEEtran}

\usepackage{graphicx}
\usepackage{amsmath}
\usepackage{array}
\usepackage{booktabs}
\usepackage{tabularx}
\usepackage{multirow}
\usepackage{hyperref}
\usepackage{cite}
\usepackage{placeins}
\usepackage{orcidlink}
\usepackage{threeparttable}  
\usepackage{makecell} 
\usepackage{textcomp}
\usepackage{makecell}
\usepackage{adjustbox}
\usepackage{multirow}
\usepackage{enumitem}
\usepackage{diagbox}
\usepackage{textcomp}
\usepackage{textcomp}
\title{On the Effectiveness of Textual Prompting with Lightweight Fine-Tuning for SAM3 Remote Sensing Segmentation}

\author{
    Roni~Blushtein-Livnon*\raisebox{0.5ex}{\orcidlink{0000-0002-3493-4894}},
    Osher~Rafaeli*\raisebox{0.5ex}{\orcidlink{0000-0002-7097-7568}},
    David~Ioffe\raisebox{0.5ex}{\orcidlink{0009-0007-5922-3558}},    
    Amir~Boger\raisebox{0.5ex}{\orcidlink{0009-0008-5072-9197}},
    Karen~Sandberg~Esquenazi\raisebox{0.5ex}{\orcidlink{0009-0001-0227-0464 }},
    and~Tal~Svoray\raisebox{0.5ex}{\orcidlink{0000-0003-2243-8532}}
    \thanks{* These authors contributed equally to this work.} \thanks{R. Blushtein-Livnon, O. Rafaeli, D. Ioffe, A. Boger, K. Sandberg Esquenazi, and T. Svoray are with the Department of Environmental, Geoinformatics and Urban Planning Sciences, Ben-Gurion University of the Negev, Israel (e-mails: livnon@bgu.ac.il;osherr@post.bgu.ac.il; ioffed@post.bgu.ac.il; bogera@post.bgu.ac.il; sandbeka@post.bgu.ac.il; tsvoray@bgu.ac.il).}
    \thanks
}

\IEEEaftertitletext{\vspace{-3\baselineskip}}

\begin{document}

\maketitle
\begin{abstract}
Remote sensing (RS) segmentation is limited by scarce annotations and domain gaps between overhead and natural imagery, motivating effective adaptation under constrained supervision. SAM3’s concept-driven framework enables mask generation from textual prompts without target-specific modification. We evaluate SAM3 across four RS targets, sources, and resolutions, comparing textual, geometric, and hybrid prompting under lightweight fine-tuning at increasing supervision levels, and zero-shot inference. Results show that combining semantic and geometric cues consistently yields the best performance, while text-only prompting performs worst, particularly for irregular targets, reflecting limited semantic alignment between textual concepts and overhead appearances. Nevertheless, lightly fine-tuned textual prompting offers a favorable performance-effort trade-off for regular targets. Performance improves sharply from zero-shot to fine-tuning, followed by diminishing returns, indicating that modest annotation effort may suffice. Persistent Precision–IoU gaps further reveal under-segmentation and boundary errors as dominant failure modes, especially for irregular and less prevalent targets.
\end{abstract}
\vspace{-3pt}
\vspace{-6pt}
\section{Introduction}
Image segmentation is fundamental to remote sensing (RS) applications but is challenged by high scene heterogeneity, irregular object boundaries, extreme scale variation, dense spatial patterns, and sensor-specific distortions  \cite{kotaridis2021remote}. Moreover, most foundation models are trained on natural images, limiting their generalization to RS data. Consequently, effective utilization of these models for RS requires both domain and target-specific adaptation \cite{10839471, gao2025combining}.

Mirroring a broader paradigm of text-promptable queries, as in LLMs and TTIs (Text-to-Image) generators, recent CV models redefined user/model interactions, shifting from static architectures to prompt-driven execution. This is evident in classifiers (e.g., CLIP \cite{radford2021learning}), object detection (e.g., Grounded DINO \cite{liu2024grounding}), and segmentation (e.g., CLIPSeg \cite{luddecke2022image}). Many adopt transformer-based vision-language models (VLM), combining ViT or Swin backbones with language encoders, yet often lack the pixel-level precision and prompting flexibility required for fine-grained RS delineation.

Recent RS studies leverage textual prompting in VLMs primarily for semantic alignment. While domain-adapted models, e.g., RemoteCLIP \cite{liu2024remoteclip}, encode meaningful semantic cues via large-scale pretraining, they rely on template-based text and do not directly support pixel-level segmentation. Others show that textual prompt formulation in RS is nontrivial, as prompts emphasizing either global scene semantics or isolated local details often fail to capture the complexity of overhead imagery \cite{sun2025strong}. These findings highlight the importance of prompt design but remain centered on cross-modal retrieval rather than segmentation \cite{zi2025visual}. Even when text is incorporated into RS semantic segmentation, as in SegCLIP \cite{zhang2024segclip}, it guides supervised, closed-vocabulary architectures rather than serving as an open, standalone interface. Overall, existing VLMs and open-vocabulary approaches focus on architectural adaptation, class-based supervision, or prompt engineering, whereas we adopt here an analytical perspective on how prompting modalities interact with RS target properties and supervision levels.

In parallel with VLMs, a complementary paradigm uses geometric cues and localized user input to guide interactive segmentation, typically via CNN-based encoder–decoder models with lightweight attention (e.g., RITM \cite{sofiiuk2022reviving}). However, adapting such geometric prompting to RS imagery is nontrivial and often requires substantial retraining \cite{10839471}.

Recently, the original SAM \cite{kirillov2023segment}, and its subsequent variant, SAM2, extended this paradigm through extremely large-scale pretraining to enable zero-shot (ZS) segmentation. However, their performance degrades on RS imagery without adaptation, particularly in limited-data settings, motivating target-aware fine-tuning (FT) \cite{chen2024rsprompter}. Although geometric prompt-based models can perform well after fine-tuning, generating high-quality prompts requires expert, labor-intensive annotation of complex scenes, especially at large scales \cite{osco2023segment}. To reduce this burden, intermediary models such as YOLO are often used to automatically generate prompts, trading annotation effort for added dependence on their accuracy and additional training.

Accordingly, hybrid prompting architectures have emerged as a bridge between VLMs and interactive segmentation. For example, SEEM \cite{zou2023segment} combines CLIP-based textual cues with spatial guidance via a shared Transformer decoder. More recently, SAM3 \cite{carion2025SAM3} introduced a concept-driven interface centered on short semantic phrases as primary prompts, unifying textual and geometric inputs within a shared embedding space processed by a ViT-based decoder. Designed to operate directly from text while optionally incorporating spatial priors, SAM3 supports seamless multimodal prompting and robust ZS segmentation, a capability that may be critical for RS tasks with limited geometric inputs. 

However, SAM3 is pretrained without overhead imagery, raising concerns about domain generalization. We address this by evaluating its RS performance across prompt strategies, target types, and training extents, analyzing when concept-driven prompting succeeds and when limited geometric supervision constrains it. Key contributions include:
    (1) Benchmarking SAM3 in RS across anthropogenic and natural targets of varying size and shape. (2) Comparing text-only, geometry-only, and hybrid prompting strategies. (3) Assessing prompting performance across aerial and satellite imagery with different spatial resolutions. (4) Evaluating performance across 3 lightweight fine-tuning scales to quantify supervision-dependent gains and the training extent required for effective adaptation, with ZS inference as a generalization baseline.
\vspace{-8pt}
\section{Materials and Methodology}
\subsection{Datasets}
\begin{figure}[h]
    \centering
    \includegraphics[width=\columnwidth]{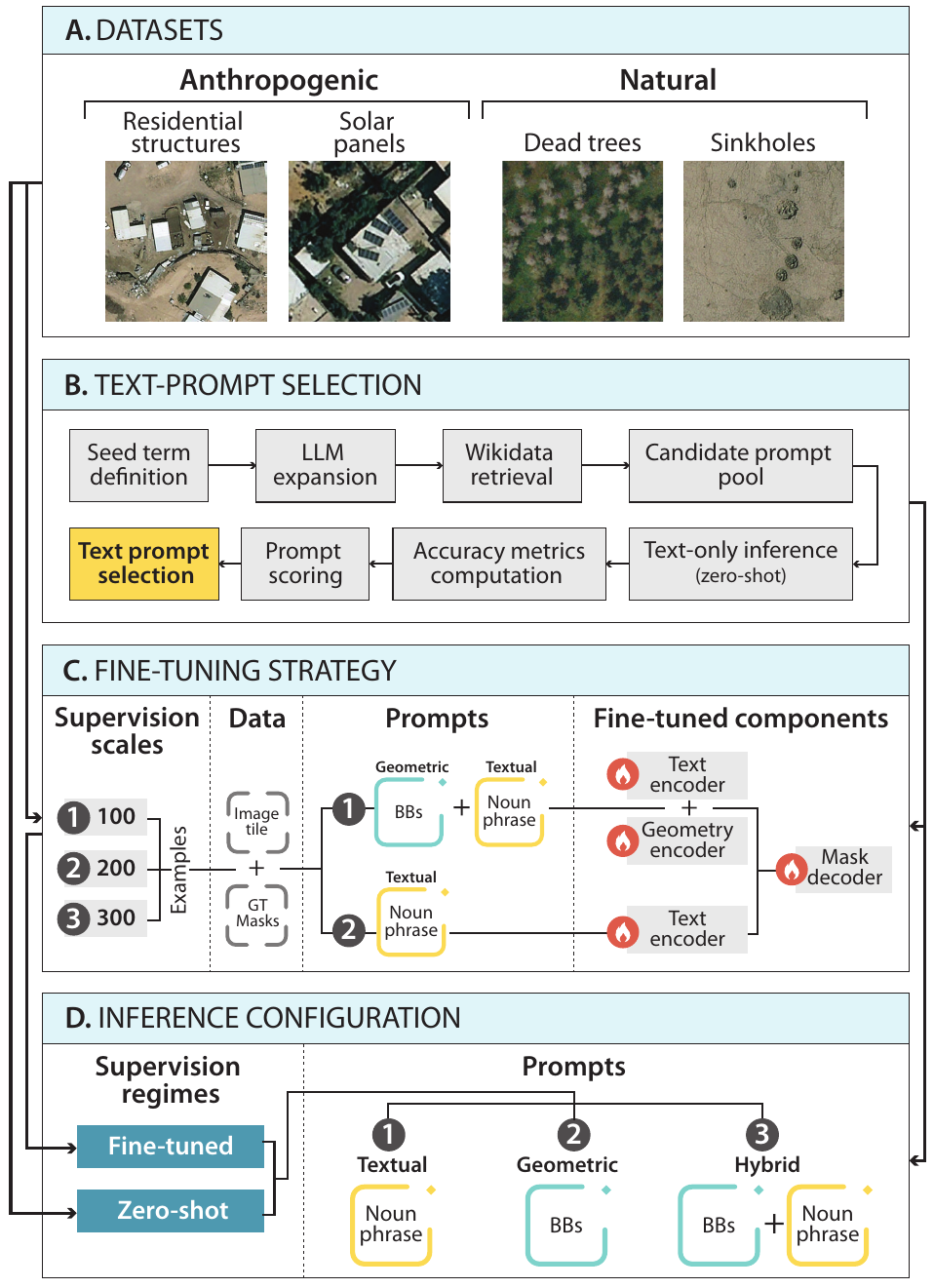}
    \vspace{-20pt}
    \caption{Experimental setup: SAM3 performance is evaluated across 4 target types using varying prompt-based training pipelines and FT scales (C). Inference relies on prompt modalities (D). Optimal conceptual text prompt for each target is determined through a structured selection procedure (B).}
    \label{framework}
\end{figure}

We constructed four independent datasets (Table \ref{data}; Fig. \ref{framework}A), each aligned with a distinct target type spanning diverse geometric properties, spatial scales, and class frequencies. The two anthropogenic targets, (1) residential structures (RSt) in rural or suburban settings and (2) small residential PVs (SP), share regular, well-defined shapes but differ in spatial extent, resulting in distinct background–target balance conditions. The two natural targets, (3) dead trees (DT) and (4) sinkholes (SH), exhibit irregular morphology, high intra-class variability, and lower real-world prevalence. 
While diverse, the selected targets do not cover the full spectrum of RS object types; thus, findings should be interpreted as indicative. For all targets, images were curated, tiled, and manually annotated by experts to produce pixel-accurate masks.
Data sources include aerial imagery with varied spatial resolutions (0.13–0.25 m/px), tiled into fixed-size patches (256 or 512 pixels) with binary masks. A cross-source comparison for anthropogenic targets contrasts aerial imagery (0.13 m/px) with WorldView satellite imagery (0.30 m/px) from distinct geographic regions.
 \vspace{-12pt}
\subsection{SAM3}
SAM3 \cite{carion2025SAM3} is a VLM for promptable, concept-level segmentation, interpreting textual prompts as abstract visual concepts to segment all corresponding object instances in an image. It unifies textual and geometric prompts via a shared Perception Encoder that projects images and text into a common embedding space. Object recognition and localization are performed by a DETR-based module with a presence head that estimates concept existence prior to localization, while pixel-level masks are generated by a MaskFormer-style decoder using learned queries rather than per-pixel classification. SAM3 reports improved ZS and fine-tuned performance over earlier versions.
\vspace{-24pt}
\subsection{Experimental Setup}
\subsubsection{Text Prompt Selection \textnormal{(Fig. \ref{framework}B)}} \label{text-selection}
SAM3 lacks mechanisms for selecting image-conditioned prompts, making prompt selection challenging. Although models, e.g., BLIP or Florence, support visual-to-text generation, their training on natural images limits their applicability to RS data, leaving prompt selection largely heuristic in overhead imagery. To address this, we introduce a refined procedure for selecting an effective textual descriptor for each target. 

Specifically, we construct a pool of domain-relevant noun phrases (up to four words, following SAM3 recommendations) from a seed term, expanded using LLM-generated variants. These are then queried against Wikidata \cite{vrandevcic2014wikidata} to ensure semantic validity and alignment with the SA-Co ontology underlying SAM3’s concept vocabulary. Let $P = \{p_1, \dots, p_M\}$ denote the resulting set of candidate prompts, where $M=10$ in our experiments. Each candidate prompt $p_j$ is evaluated on a held-out validation set consisting of 50 annotated image-mask pairs by applying SAM3 in a text-only prompting mode. A scalar score $S(p_j)$ is computed from average F1 and IoU scores and used to rank candidate prompts. The selected prompt $p_{\text{opt}}$ is: 
\vspace{-3pt}
\[
p_{\text{opt}} = \arg\max_{p_j \in P} S(p_j).
\]
\vspace{-4pt}
\subsubsection{FT Strategy \textnormal{(Fig. \ref{framework}C)}}
Each target is fine-tuned with 100, 200, and 300 objects, corresponding to supervision scales FT100, FT200, and FT300. The training data consists of image tiles paired with ground-truth binary masks. Tiles were randomly sampled from annotated imagery across different scenes and acquisition conditions, and split into training, validation, and test sets. The validation set size was matched to the corresponding training set. FT is conducted using two complementary prompting pipelines: a textual-only pipeline based on target-specific noun phrases (see Section~\ref{text-selection}) and a geometric-textual pipeline that augments textual prompts with BBs. In the textual-only pipeline, the text encoder is unfrozen to adapt semantic representations to RS, while all visual and geometric backbones remain fixed. In the geometric-textual pipeline, spatial guidance is provided via automatically generated BBs derived from ground-truth binary masks, and both text and geometry encoders are unfrozen to enable joint semantic and spatial conditioning. In both, the mask decoder is unfrozen, whereas the vision encoder and DETR encoder-decoder are kept frozen to preserve pretrained representations. All FT runs use identical hyperparameters (LR=$1e^{-5}$, 25 epochs, batch size=2) to ensure comparability across supervision scales and target types.
\vspace{2pt}
\subsubsection{Inference Configuration \textnormal{(Fig. \ref{framework}D)}} Preliminary experiments indicate that performance degrades when training and inference prompt configurations are mismatched. This is likely due to increased reliance on explicit spatial cues during FT, which may limit the model’s ability to localize target instances based solely on textual semantics during inference. Consequently, inference is performed using prompt configurations that are consistent with the corresponding FT setup.\\
Inference was performed on 500 held-out samples per target under three prompting modes: (1) text-only prompting using selected noun phrases; (2) geometric-only prompting based on BBs; and (3) hybrid prompting that combines geometric and textual cues. ZS segmentation is evaluated in parallel under similar prompting configurations, using pretrained models without task-specific adaptation. In addition, SAM3 was compared with a mainstream CNN model, U-Net, using an ImageNet-pretrained ResNet-50 encoder.
\vspace{2pt}
\subsubsection{Evaluation Strategy}
Segmentation performance is evaluated along four dimensions. These include: ZS versus FT; supervision scale effect; differences in prompting configuration; and comparisons across targets. Evaluation metrics include: precision, recall, F1, and IoU. An additional cross-source (aerial, satellite) comparison was conducted for RSt and SP. Per-metric differences were computed by subtracting the best-performing satellite result from the best-performing aerial result.

\begin{table}[t]
\centering
\caption{Dataset and target characteristics}
\vspace{-6pt}
\label{tab:dataset_summary}
\setlength{\tabcolsep}{4pt}
\renewcommand{\arraystretch}{0.95}
\begin{tabular}{lccl}
\toprule
\textbf{Target} & \textbf{\# Tiles} & \textbf{Size$^*$ (\scriptsize m$^2$)}& {\textbf{Location}} \\
\midrule
RSt$^{**}$  & 790 & 56.2 (\scriptsize 77.3) & Northern Negev, Israel; CA, US \\
SP$^{**}$  & 975 & 8.3 (\scriptsize 6.3) &  Northern Negev, Israel; CA, US \\
DT & 250 & 3.7 (\scriptsize 7.2) & Lahav Forest, Israel \\
SH & 330 & 70.9 (\scriptsize 149.3) & Dead Sea, Israel\\
\bottomrule
\end{tabular}

\vspace{2pt}
\footnotesize
\textit{Notes: $^{*}$Mean (\scriptsize SD); $^{**}$Total tiles (aerial and satellite).}
 \label{data}
\end{table}

\vspace{-15pt}
\section{Results}
Prompts were selected based on ZS performance across 10 semantically varying noun phrases per target. As an illustrative example, for SP, candidates such as \textit{tilted solar panels}, \textit{residential solar panels}, \textit{photovoltaic installation}, and \textit{rooftop and ground PV} were evaluated, with \textit{tilted solar panels} achieving the highest scores (IoU 68.7\%, F1 81.4\%). Across targets, generic or abstract prompts (e.g., \textit{soil depressions}, \textit{tree mortality}) and those reflecting ground-level semantics (e.g., \textit{standing dead tree}, \textit{concave terrain anomaly}) tended to underperform. In contrast, visually descriptive noun phrases aligned with RS appearance consistently achieved superior performance. Accordingly, the selected prompts were \textit{rectangular structure} (RSt), \textit{tilted solar panels} (SP), \textit{healthy woody vegetation trees} (LT), \textit{silvery dead crown aerial view} (DT), and \textit{dark circular void in soil} (SH).

\begin{figure}[h!]
    \centering
    \includegraphics[width=\columnwidth]{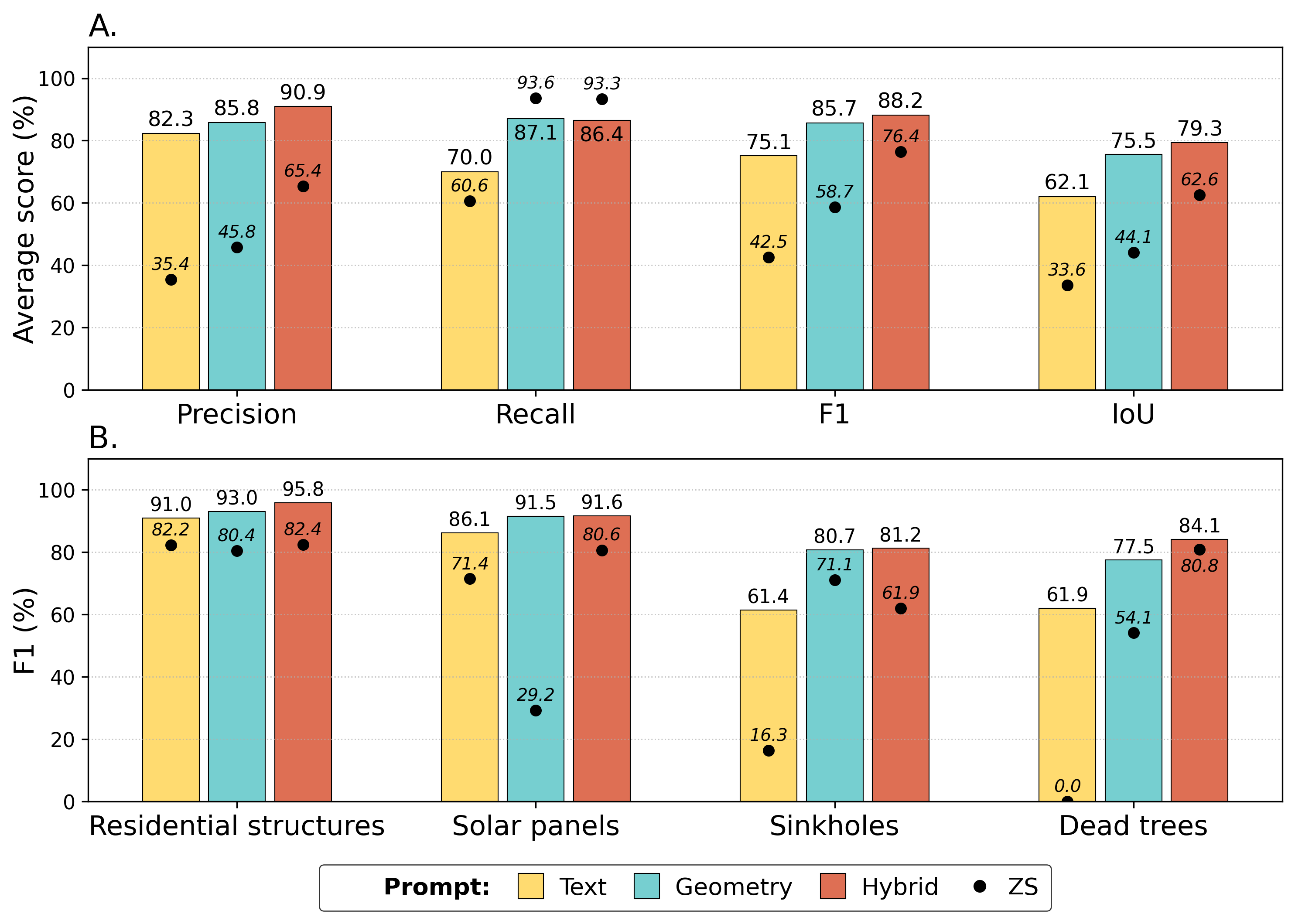}
    \vspace{-20pt}
    \caption{Performance under FT200 and ZS by prompt type. A: Average metric scores across all targets; B: F1 scores by target.}
    \label{average}    
\end{figure} 

The experimental results are presented in Fig. \ref{average}, Table \ref{table}, and Table \ref{H_L_res}. As shown in Fig. \ref{average}A, the hybrid prompt achieves the highest performance across all metrics (averaged overall targets), with the exception of a small advantage of the geometric prompt in Recall. Text prompt consistently yields the lowest scores across all metrics and targets (Fig. \ref{average}A-B). FT200 improves all metrics relative to ZS, except for Recall, with the largest gains observed in precision. Improvements by the hybrid prompt remain modest, given its already high ZS scores. Clear differences emerge between Precision and IoU, with IoU scores 10\% or lower, indicating a systematic reduction in geometric accuracy across targets. 

\newcommand{\MetricFont}{\fontsize{1.9}{3}\selectfont}       
\newcommand{\ScaleFont}{\fontsize{1.9}{3}\selectfont}        
\newcommand{\PromptCellFont}{\fontsize{1.7}{3}\selectfont}
\newcommand{\PromptCellGap}{-2.9pt}
\newcommand{\RotPromptCell}[1]{%
  \rotatebox{90}{%
    {\PromptCellFont
      \shortstack[c]{\textbf{#1}\\[\PromptCellGap]{prompt}}%
    }%
  }%
}
\newcommand{\TargetHead}{\shortstack[c]{\textbf{Tar-}\\[-2.5pt]\textbf{get}}}

\begin{table*}[!htbp]
\centering
\setlength{\arrayrulewidth}{0.1pt}
\setlength{\abovecaptionskip}{2pt}
\caption{Performance metrics (\%) across fine-tuning scales by prompt and target types.}
\vspace{-3pt}
\begin{threeparttable}
\begin{adjustbox}{width=\textwidth}
\fontsize{1.7}{2.0}\selectfont
\setlength{\tabcolsep}{1.1pt}
\renewcommand{\arraystretch}{0.95}
\begin{tabular}{c|>{\raggedright\arraybackslash}p{2.0em}|cccc|cccc|cccc|cccc}
\hline
& \multirow{2}{*}{\TargetHead} &
\multicolumn{4}{c|}{{\MetricFont\textbf{Precision}}} &
\multicolumn{4}{c|}{{\MetricFont\textbf{Recall}}} &
\multicolumn{4}{c|}{{\MetricFont\textbf{F1}}} &
\multicolumn{4}{c}{{\MetricFont\textbf{IoU}}} \\
& &
\multicolumn{1}{c}{{\ScaleFont{ZS}}} &
{\ScaleFont{FT100}} &
{\ScaleFont{FT200}} &
{\ScaleFont{FT300}} &
\multicolumn{1}{c}{{\ScaleFont{ZS}}} &
{\ScaleFont{FT100}} &
{\ScaleFont{FT200}} &
{\ScaleFont{FT300}} &
\multicolumn{1}{c}{{\ScaleFont{ZS}}} &
{\ScaleFont{FT100}} &
{\ScaleFont{FT200}} &
{\ScaleFont{FT300}} &
\multicolumn{1}{c}{{\ScaleFont{ZS}}} &
{\ScaleFont{FT100}} &
{\ScaleFont{FT200}} &
{\ScaleFont{FT300}} \\
\hline
\multirow{4}{*}{\RotPromptCell{Text}} & RSt
& 73.18 & 90.40 & \textbf{92.62} & 91.93
& 93.88 & 90.95 & 89.38 & \textbf{91.01}
& 82.25 & 90.67 & 90.97 & \textbf{91.47}
& 69.85 & 82.94 & 83.44 & \textbf{84.28} \\
& SH
& 9.50 & 81.14 & 79.49 & \textbf{88.81}
& 58.42 & 44.08 & \textbf{50.06} & 39.42
& 16.34 & 57.13 & \textbf{61.44} & 54.64
& 8.90 & 39.99 & \textbf{44.34} & 37.56 \\
& DT
& 0.00 & 66.78 & 70.11 & \textbf{77.51}
& 0.00 & 51.48 & \textbf{55.45} & 47.08
& 0.00 & 58.14 & \textbf{61.93} & 58.58
& 0.00 & 40.98 & \textbf{44.85} & 41.42 \\
& SP
& 59.10 & 88.60 & 87.00 & \textbf{92.14}
& \textbf{90.29} & 78.97 & 85.28 & 79.61
& 71.44 & 83.51 & \textbf{86.13} & 85.42
& 55.57 & 78.97 & \textbf{75.64} & 74.54 \\
\hline
\multirow{4}{*}{\RotPromptCell{Geometry}} & RSt
& 67.60 & 84.82 & 88.68 & \textbf{94.09}
& \textbf{99.0} & 97.39 & 97.77 & 94.90
& 80.40 & 90.67 & 93.00 & \textbf{94.49}
& 67.20 & 82.93 & 86.92 & \textbf{89.56} \\
& SH
& 59.90 & 72.19 & \textbf{95.54} & 77.37
& 87.50 & 84.25 & 69.86 & \textbf{91.52}
& 71.11 & 77.76 & 80.71 & \textbf{83.86}
& 55.20 & 63.61 & 67.66 & \textbf{72.20} \\
& DT
& 38.39 & \textbf{72.27} & 70.51 & 55.62
& 91.72 & 89.18 & 85.92 & \textbf{92.63}
& 54.13 & \textbf{79.84} & 77.46 & 69.50
& 37.10 & \textbf{66.44} & 63.21 & 53.26 \\
& SP
& 17.19 & 85.11 & 88.55 & \textbf{92.48}
& \textbf{96.33} & 94.86 & 94.70 & 92.89
& 29.17 & 89.72 & 91.52 & \textbf{92.69}
& 17.08 & 81.36 & 84.37 & \textbf{86.37} \\
\hline
\multirow{4}{*}{\RotPromptCell{ResUNet}} & 
RSt
& -- & 88.50 & 91.74 & 62.11
& -- & 87.98 & 88.12 & 95.78
& -- & 88.24 & 89.90 & 75.35
& -- & 78.96 & 81.65 & 60.46 \\

& PV
& -- & 75.34 & 80.26 & 87.20
& -- & 71.89 & 76.88 & 77.78
& -- & 73.57 & 78.53 & 82.22
& -- & 58.20 & 64.65 & 69.81 \\

& SH
& -- & 39.40 & 62.30 & 37.16
& -- & 53.25 & 33.26 & 71.58
& -- & 45.29 & 43.37 & 48.93
& -- & 29.27 & 27.69 & 32.39 \\

& SP
& -- & 70.86 & 75.72 & 85.39
& -- & 30.44 & 41.10 & 41.00
& -- & 42.59 & 53.28 & 55.40
& -- & 27.06 & 36.31 & 38.31 \\
\hline
\multirow{4}{*}{\RotPromptCell{Hybrid}} & RSt
& 71.07 & 96.07 & 94.00 & \textbf{98.22}
& \textbf{98.18} & 95.80 & 97.64 & 92.36
& 82.40 & \textbf{95.94} & 95.79 & 95.20
& 70.10 & \textbf{92.19} & 91.91 & 90.84 \\
& SH
& 47.32 & 81.83 & \textbf{96.26} & 88.28
& \textbf{89.60} & 84.96 & 70.22 & 89.35
& 61.93 & 83.37 & 81.20 & \textbf{88.81}
& 44.86 & 71.48 & 68.35 & \textbf{79.87} \\
& DT
& 73.68 & 76.20 & \textbf{84.02} & 77.91
& 89.49 & 88.51 & 84.09 & \textbf{90.27}
& 80.82 & 81.89 & \textbf{84.06} & 83.64
& 67.81 & 69.34 & \textbf{72.50} & 71.88 \\
& SP
& 69.43 & 87.51 & 89.49 & \textbf{93.34}
& \textbf{96.12} & 94.29 & 93.83 & 92.38
& 80.62 & 90.77 & 91.61 & \textbf{92.86}
& 67.54 & 83.10 & 84.51 & \textbf{86.67} \\
\hline
\end{tabular}
\end{adjustbox}
\begin{tablenotes}[flushleft]
\small
\begin{minipage}{\textwidth}
\item \footnotesize{\textit{Abbreviations: RSt - Residential Structures; SP - Solar Panels; SH - Sinkholes; DT - Dead Trees. Bold values denote the highest score.}} 
\end{minipage}
\end{tablenotes}
\label{table}
\end{threeparttable}
\end{table*}

\begin{table}[t]
\centering
\caption{Per-metric differences (\%) between the best-performing results obtained from aerial and satellite imagery.}
\vspace{-7pt}
\label{H_L_res}
\setlength{\tabcolsep}{6pt}
\fontsize{7.9}{7.5}\selectfont
\renewcommand{\arraystretch}{0.75}
\begin{tabular*}{0.9\linewidth}{@{\extracolsep{\fill}}llrrrr}
\toprule
\textbf{Prompt} & \textbf{Object} & \textbf{Precision} & \textbf{Recall} & \textbf{F1} & \textbf{IoU} \\
\midrule
\multirow{2}{*}{\textbf{Text}} 
 & RSt & 5.1 & 16.3 & 9.2 & 14.5 \\
 & SP  & 11.6 & 16.3 & 9.5 & 17.9 \\
\addlinespace[1pt]
\hline
\addlinespace[1pt]
\multirow{2}{*}{\textbf{Geometry}} 
 & RSt & 10.1 & 6.7 & 8.7 & 3.7 \\
 & SP  & 1.0 & 3.8 & 1.2 & -7.5 \\
\addlinespace[1pt]
\hline
\addlinespace[1pt]
\multirow{2}{*}{\textbf{Hybrid}} 
 & RSt & 12.7 & 8.3 & 8.8 & 15.0 \\
 & SP  & 1.3 & 3.9 & 1.4 & 2.3 \\
\bottomrule
\end{tabular*}
\end{table}

Fig. \ref{average}B further reveals pronounced and consistent differences between target categories. Anthropogenic, regularly shaped targets consistently achieve substantially higher scores than natural, irregularly shaped targets. RSt achieves high scores already under ZS across all prompts, and this advantage persists after FT200. Under text prompting, performance gaps between target categories widen substantially, reaching up to thirty percent after FT200. This disparity is even more pronounced under ZS, where performance drops sharply for natural targets and fails entirely for DT.
Table \ref{table} illustrates the model’s learning score and trajectory across supervision scales and compares its performance with ResUNet \cite{DIAKOGIANNIS202094}. Averaged across prompting strategies, the largest performance gain occurs between ZS and FT100 for all targets (F1 gain increases from $\sim$10.7\% for RSt up to $\sim$28.3\% for DT), while further training yields partial or full metric convergence. Differences between FT200 and FT300 are negligible and occasionally negative (F1 change: $\sim$$-$3.9\% for DT up to $\sim$$+$1.3\% for SH). The hybrid prompt is the most stable across metrics and targets, showing only moderate improvements with increasing supervision due to its relatively strong ZS performance. In contrast, the text prompt consistently lags behind other prompts, even after FT300. Except for RSt, its Recall score declines at FT300, and for several targets, its geometric accuracy (IoU) also deteriorates. Anthropogenic, regularly shaped, targets achieve substantially higher performance from the FT100 onward, with small variability across metrics, whereas natural, irregularly shaped targets display lower scores and greater fluctuations across supervision scales. ResUNet achieves reasonable performance after FT on 300 samples, particularly on RSt and PV (IoU of 60.46\% and 69.81\%, respectively); however, it consistently underperforms all SAM3 prompting configurations.
\vspace{2.5pt}

Table \ref{H_L_res} reports performance differences across RS sources, contrasting aerial and WorldView satellite imagery at different spatial resolutions. This comparison is restricted to the anthropogenic targets, for which comparable cross-source datasets were available. A consistent advantage of aerial imagery is observed, most pronounced under text prompting, with large gains in Recall and IoU for both targets. Geometry prompting substantially reduces cross-source gaps and, for SP, even reverses the IoU advantage, indicating that spatial cues mitigate resolution-related limitations. Hybrid prompting combines robustness and accuracy, yielding clear gains for RSt while largely equalizing performance across sources for SP. Overall, cross-source sensitivity is highest for text-only prompting and lowest when geometric information is incorporated.
\vspace{-6pt}
\section{Discussion and Conclusion}
Our results reveal consistent performance differences across prompt strategies, target categories, supervision scales, and RS sources. The hybrid prompt achieves the highest overall accuracy across metrics, indicating that combining semantic cues with explicit spatial constraints supports stable mask generation. This behavior is consistent with a complementary conditioning mechanism, where textual prompts provide coarse semantic filtering of candidate regions, while geometric prompts impose explicit spatial priors that reduce localization ambiguity during mask decoding. In contrast, geometric prompting alone tends to produce more inclusive predictions, yielding slightly higher Recall at the expense of boundary accuracy. Text-only prompting consistently underperforms, reflecting a limited semantic alignment between SAM3’s text embeddings and RS visual representations, particularly for natural targets whose appearance in overhead imagery diverges from ground-level views \cite{deng2025dual}, with DT representing an extreme case of visual-semantic embedding mismatch. From a representation-learning perspective, this divergence translates into a mismatch between the visual cues encoded by the model’s textual embeddings, learned from object-centric textures and viewpoints, and the reflectance-based, scale-compressed, and planimetric representations that characterize RS imagery, thereby weakening cross-modal correspondence \cite{sun2025strong}. RSt constitute a notable exception, likely due to the target prevalence as a semantically well-defined concept in the model corpora, as well as its large spatial extent, regular geometry, and high contrast with the background, which reduces segmentation ambiguity.

The cross-source comparison further indicates that prompting strategies also differ in their sensitivity to sensing modality and resolution. Text-only prompting is the most source-dependent, with substantially larger gains for aerial imagery, particularly in Recall and IoU, reflecting its reliance on fine-grained visual detail. Geometric prompting markedly reduces cross-source disparities, suggesting that explicit spatial cues partially compensate for reduced image detail. Hybrid prompting exhibits the most robust cross-source behavior, preserving the aerial advantage for larger, regular targets while largely equalizing performance for smaller objects. The smaller cross-source gaps observed for SP may reflect their high visual consistency across locations, in contrast to RSt, whose appearance and surrounding context vary more strongly across regions.

Across targets, a systematic gap between Precision and IoU indicates that SAM3 generally identifies object presence correctly but struggles with accurate boundary delineation. Under-segmentation emerges as the dominant error mode, characterized by frequent FNs and shape distortions along boundaries. These effects are most pronounced for natural targets with fine-scale geometric variability and, in the case of DT, weak spectral contrast, confirming that geometric reasoning remains a primary source of challenge even when most target pixels are correctly identified.

Target-specific differences further highlight these trends. Anthropogenic targets consistently achieve higher scores across metrics, including under text prompting, and exhibit strong ZS performance with low variability across supervision scales. This likely indicates stronger inductive alignment with  visual and semantic priors encoded by SAM3, whose favorable geometric characteristics enable reliable boundary delineation. SP, however, exhibits a more complex response. Under ZS, performance is high when semantic cues are available but degrades under purely geometric prompting, likely because the angular positioning of small PVs increases the background-to-target ratio within BBs. Limited FT reverses this trend, with geometric constraints becoming a primary driver of performance gains. Compared with RSt, SP lower performance is plausibly explained by differences in target size and background contrast. In our datasets, PVs appear on  rooftops and the ground, the latter exhibiting low contrast with its surroundings. In contrast, natural targets consistently achieve lower segmentation performance. Within this category, SH exhibits comparatively improved boundary inference, likely due to internal, depth-induced shadowing that enhances object-background contrast, whereas DT remains low due to reduced spectral contrast and fragmented geometry. Accordingly, prompting effectiveness in RS can be interpreted as target-dependent: geometrically regular and spectrally distinctive objects benefit more from semantic cues, whereas irregular, fragmented, or low-contrast targets require explicit spatial constraints to compensate for weaker semantic embeddings.
Finally, improvements across supervision scales exhibit clear diminishing returns. Most gains occur between ZS inference and FT100, with only marginal improvements at FT200 and FT300. As supervision increases, performance stabilizes, while the relative ranking of prompt strategies and target categories is preserved, indicating that limited target-specific data are sufficient for effective adaptation to RS. Still, text prompting exhibits a slight decline in Recall at FT300, suggesting reduced generalization when textual cues become tightly coupled to a narrow fine-tuning distribution.
In sum, our results support four conclusions: (1) Hybrid prompting is recommended for RS segmentation, particularly under limited supervision, as it consistently yielded the highest performance across targets and metrics. (2) Text-only prompting should be used with caution. While generally insufficient for accurate delineation, it can provide a favorable performance-to-effort trade-off for targets that are visually salient, geometrically simple, and well represented in the model’s pre-training data, where short FT regimes may yield meaningful gains without the need for geometric annotation. (3) Segmentation performance is strongly target dependent and sensitive to sensing conditions. Object geometry and visual consistency across RS sources, as well as their representation in pre-training data, critically influence model performance. (4) Limited FT of SAM3 is adequate for adaptation. Most gains are achieved with a small dataset, while additional supervision primarily stabilizes predictions, indicating that modest annotation effort is sufficient for practical deployment.
 \vspace{-6pt}
\section*{Acknowledgment}
\vspace{-4pt}
This work was supported by the Israel Science Foundation (ISF), Grant 299/23.
\vspace{-10pt}
\bibliographystyle{ieeetr} 
 \bibliography{references}
\end{document}